\documentclass[conference]{IEEEtran}
\IEEEoverridecommandlockouts
\usepackage{cite}
\usepackage[font=small,labelfont=bf]{caption}
\usepackage{pgfplots}
\pgfplotsset{compat=1.17} %
\usepackage{amsmath,amssymb,amsfonts}
\usepackage{algorithmic}
\usepackage{graphicx}
\usepackage{textcomp}
\usepackage{booktabs}  
\usepackage{multirow}  
\usepackage{amsmath}   
\usepackage{xcolor}
\def\BibTeX{{\rm B\kern-.05em{\sc i\kern-.025em b}\kern-.08em
    T\kern-.1667em\lower.7ex\hbox{E}\kern-.125emX}}
\usepackage{graphicx}   
\usepackage{subcaption} 
\usepackage{float}      
    
\begin{document}

\title{Enhancing Fingerprint Recognition Systems: Comparative Analysis of Biometric Authentication Algorithms and Techniques for Improved Accuracy and Reliability \\
}


\author{
     Temirlan Meiramkhanov\IEEEauthorrefmark{1}, Arailym Tleubayeva\IEEEauthorrefmark{2}, \\

    \IEEEauthorblockA{\IEEEauthorrefmark{1}Department of Computational and Data Sciences, Astana IT University, Astana, Kazakhstan \\
    Email: 242813@astanait.edu.kz}
    \IEEEauthorblockA{\IEEEauthorrefmark{2}Department of ComputerEngineeringAstana IT University, Astana, Kazakhstan \\ 
    Email: a.tleubayeva@astanait.edu.kz}

}

\maketitle

\begin{abstract}

Fingerprint recognition systems stand as pillars in
the realm of biometric authentication, providing indispensable
security measures across various domains. This study investi-
gates integrating Convolutional Neural Networks (CNNs) with
Gabor filters to improve fingerprint recognition accuracy and
robustness. Leveraging a diverse dataset sourced from the Sokoto
Coventry Fingerprint Dataset, our experiments meticulously
evaluate the efficacy of different classification algorithms.
Our findings underscore the supremacy of CNN-based ap-
proaches, boasting an impressive overall accuracy of 94\%.
Furthermore, the amalgamation of Gabor filters with CNN
architectures unveils promising strides in discerning altered
fingerprints, illuminating new pathways for enhancing biometric
authentication systems.
While the CNN-Gabor fusion showcases commendable perfor-
mance, our exploration of hybrid approaches combining mul-
tiple classifiers reveals nuanced outcomes. Despite these mixed
results, our study illuminates the transformative potential of deep
learning methodologies in reshaping the landscape of fingerprint
recognition.
Through rigorous experimentation and insightful analysis,
this research not only contributes to advancing biometric au-
thentication technologies but also sheds light on the intricate
interplay between traditional feature extraction methods and
cutting-edge deep learning architectures. These findings offer
actionable insights for optimizing fingerprint recognition systems
for real-world deployment, paving the way for enhanced security
and reliability in diverse applications.

\end{abstract}


\section{INTRODUCTION}
Fingerprint recognition systems have established themselves as cornerstones of biometric authentication, providing unparalleled reliability and accuracy across diverse applications in security, law enforcement, and access control \cite{yang2019security}. Biometric authentication is a technology that verifies identity based on unique biological characteristics \cite{yang2019security}.

Deep learning and advanced image processing have significantly enhanced these systems' performance and robustness \cite{zeng2019research}. Deep learning is a subset of artificial intelligence that has demonstrated remarkable success in various tasks \cite{zeng2019research}.

Traditionally, fingerprint recognition systems have relied on hand-crafted features such as minutiae points  \cite{valdes2019review}. These are specific, predefined characteristics that are manually extracted from fingerprint images for the purpose of identification \cite{valdes2019review}. However, the performance of these methods can degrade with variations in fingerprint quality \cite{medina2012improving}.

In recent years, Convolutional Neural Networks (CNNs), a type of deep learning, have demonstrated remarkable success in image processing tasks \cite{ekpo2019modelling}\cite{zeng2019research}. CNNs can automatically learn discriminative features from raw image data, offering the potential to outperform traditional feature extraction techniques for fingerprint recognition \cite{zeng2019research}.

Furthermore, Gabor filters, known for their texture analysis capabilities \cite{pradeep2022revolutionary}\cite{praseetha2019secure}, can potentially provide complementary information to enhance deep learning-based fingerprint recognition systems \cite{pradeep2022revolutionary}. Gabor filters are linear filters used in image processing for texture analysis, which means they can capture and highlight the unique patterns in a fingerprint image \cite{pradeep2022revolutionary}\cite{praseetha2019secure}.

This research endeavors to investigate the integration of Convolutional Neural Networks (CNNs) with Gabor filters for enhanced fingerprint recognition \cite{pradeep2022revolutionary}\cite{medina2012improving}. By capitalizing on the intricate texture details within fingerprint images, we aspire to leverage the complementary advantages of deep learning models and established feature extraction methodologies \cite{praseetha2019secure}\cite{patel2019improved}.

Through comprehensive analysis and experimentation, we strive to evaluate the effectiveness of various classification algorithms and assess their potential for real-world implementation \cite{patel2019improved}. Our study aims to enrich the domain of fingerprint recognition by proposing a distinctive framework that unites deep learning and traditional feature extraction techniques, ultimately contributing to advancements in the realm of biometric authentication \cite{yang2019security}\cite{priesnitz2022mobile}.

The primary objective of this research is to evaluate the integration of Convolutional Neural Networks (CNNs) with Gabor filters for enhanced fingerprint recognition \cite{pradeep2022revolutionary}\cite{medina2012improving}. By investigating the performance of various classification algorithms and assessing their potential for real-world implementation, we aim to enrich the domain of fingerprint recognition and contribute to advancements in biometric authentication technology \cite{yang2019security}\cite{priesnitz2022mobile}.
\section{Literature review}

Fingerprint recognition, a vital component of biometric authentication systems, continues to evolve with advancements in algorithms and methodologies. This literature review synthesizes key findings from recent studies in the field of fingerprint recognition, providing insights into various approaches and their implications for system accuracy and efficiency.

Several studies have investigated minutia descriptors' effectiveness in fingerprint recognition. \cite{ekpo2019modelling} Valdes-Ramirez et al. (2019) conducted a detailed evaluation of different descriptors, highlighting the importance of considering factors such as fingerprint rotation in algorithm performance. This research contributes to understanding the nuances of feature representations crucial for accurate recognition.

Algorithm development plays a critical role in enhancing fingerprint recognition systems. \cite{ekpo2019modelling} Ekpo et al. (2019) proposed a novel algorithm focusing on minutiae extraction to address common challenges such as poor image quality and orientation variations. Their approach offers practical insights into improving system robustness and efficiency, essential for real-world applications.

Deep learning methodologies have emerged as promising approaches in fingerprint recognition research. \cite{zeng2019research} Zeng et al. (2018) explored deep learning for partial fingerprint recognition, showcasing superior performance compared to traditional methods. While not all aspects of their methodology may directly align with every project, their findings offer valuable insights into the potential of deep learning in fingerprint recognition.

Hybrid verification systems combining statistical and neural network-based approaches have been proposed to enhance system security and accuracy. \cite{praseetha2019secure} Praseetha et al. (2019) advocated for such a system, leveraging Convolutional Neural Networks (CNNs) for pre-verification. Their research underscores the importance of integrating diverse techniques to achieve robust fingerprint recognition systems.

These studies contribute to the ongoing discourse on improving fingerprint recognition algorithms' accuracy and efficiency. By synthesizing findings from diverse methodologies, researchers can gain valuable insights into potential avenues for advancing biometric authentication systems.
\section{METHODOLOGY}

\textbf{1. Dataset Description:}
We used the Sokoto Coventry Fingerprint Dataset from Kaggle \cite{shehu2018sokoto}\cite{papi2016generation}\cite{shehu2018detection}, featuring various fingerprint images under different conditions, ideal for evaluating recognition algorithms.

\textbf{2. Data Preprocessing:}
The fingerprint images were subjected to preprocessing steps to normalize and standardize their size for consistent model input. The preprocessing steps varied slightly depending on the requirements of each experiment, which are detailed in their respective subsections below.

\textbf{3. Feature Extraction:}
For certain experiments, Gabor filters were applied to the preprocessed images to extract detailed textural features at multiple orientations and scales, thereby capturing the unique textural patterns present in fingerprints.

\textbf{4. Model Architecture:}
\begin{itemize}
\item In the CNN-based experiments, a Convolutional Neural Network architecture with three convolutional layers was employed, each followed by max-pooling layers to extract features, culminating in fully connected layers and a softmax output layer for classification.
\item For experiments involving Logistic Regression and the K-Nearest Neighbors classifier, features extracted were subjected to Principal Component Analysis (PCA) for dimensionality reduction, facilitating efficient training and classification.
\item Training of CNN models was conducted using the Adam optimizer, with a learning rate set to 0.001, a batch size of 32, and a predefined number of epochs to achieve model convergence, with cross-entropy loss to optimize the model parameters.
\end{itemize}

\textbf{5. Evaluation Metrics:}
Model performance was quantified using accuracy, precision, recall, and F1-score, providing a comprehensive assessment of each model’s classification ability.

\textbf{6. Dataset Split:}
The dataset was partitioned into training and testing sets with an 80/20 ratio to ensure an adequate volume of data for training the models while retaining a substantial portion for testing and validation.

\textbf{Experiment-Specific Methodology Details:}
\begin{itemize}
\item Experiment 1 (CNN for Fingerprint Recognition): Images were resized to 32x32 pixels before being input into the CNN, which consisted of three convolutional layers.

\item Experiment 2 (CNN + Gabor Filters): Images were resized to 16x16 pixels. Gabor filters with predefined orientations and scales were applied before input into the CNN for classification.

\item Experiment 3 (Logistic Regression with PCA and SMOTE): Preprocessing included resizing images to 16x16 pixels and applying PCA for dimensionality reduction. The Synthetic Minority Over-sampling Technique (SMOTE) was used to mitigate class imbalance during training.

\item Experiment 4 (KNN Classifier): After preprocessing and applying Gabor filters, the extracted features were used to train the K-Nearest Neighbors classifier.

\item Experiment 5 (Hybrid MLP + SVM): This experiment employed a hybrid approach that involved training both Multi-Layer Perceptron (MLP) and Support Vector Machine (SVM) classifiers on the features extracted using the methodology described above.
\end{itemize}

\section{RESULTS}
In this section, we present the outcomes of our experiments aimed at enhancing fingerprint recognition using a combination of Convolutional Neural Networks (CNNs) and Gabor filters. Our investigation focused on evaluating the effectiveness of different classification algorithms in recognizing fingerprint patterns under various conditions, including alterations in image quality and difficulty levels. The experiments aimed to address the limitations of traditional fingerprint recognition methods and explore the potential of deep learning techniques in improving accuracy and robustness.

Overall, our results demonstrate promising advancements in fingerprint recognition performance. We observed substantial improvements in accuracy, precision, and recall metrics across different experimental setups. Additionally, our comparative analysis of CNN-based approaches and traditional feature extraction methods highlights the advantages of leveraging deep learning architectures for fingerprint recognition tasks. Furthermore, the integration of Gabor filters with CNNs proved to be an effective strategy for enhancing feature representation and classification accuracy.

In the following sections, we provide a detailed overview of each experiment, including the experimental setup, results analysis, and comparative insights.
\subsection*{Experiment 1: CNN-Based Fingerprint Recognition}
\subsubsection*{A. Setup Description}
Experiment 1 utilized a CNN architecture for fingerprint recognition. We processed images from four categories: real and three altered levels—easy, medium, and hard. The preprocessing involved resizing images to 32x32 pixels and normalizing to ensure uniform input. The CNN comprised three convolutional layers with ReLU activations, followed by max-pooling layers. The first convolutional layer had 32 (3x3) filters; the second had 64 (3x3) filters. To mitigate overfitting, a dropout layer with a 0.5 rate was included. Classification was performed using a softmax layer. Training utilized the Adam optimizer with a 0.001 learning rate and categorical cross-entropy loss.
\subsubsection*{B. Results Presentation:}
The CNN-based method achieved high accuracy and balanced precision and recall across all categories, as detailed the table below (TABLE I):
\begin{table}[htbp]
    \centering
    \caption{Classification metrics for experiment 1.}
    \begin{tabular}{|l|c|c|c|c|}
        \hline
        \textbf{Experiment} & \textbf{Accuracy} & \textbf{Precision} & \textbf{Recall} & \textbf{F1-Score} \\
        \hline
        Real    & 0.94 & 0.87 & 0.86 & 0.87 \\
        Easy    & 0.94 & 0.94 & 0.93 & 0.94 \\
        Medium  & 0.94 & 0.96 & 0.95 & 0.95 \\
        Hard    & 0.94 & 0.96 & 0.98 & 0.97 \\
        \hline
    \end{tabular}
    \label{tab:classification_metrics}
\end{table}
\begin{figure}[htbp]
    \centering
    \includegraphics[width=0.5\textwidth]{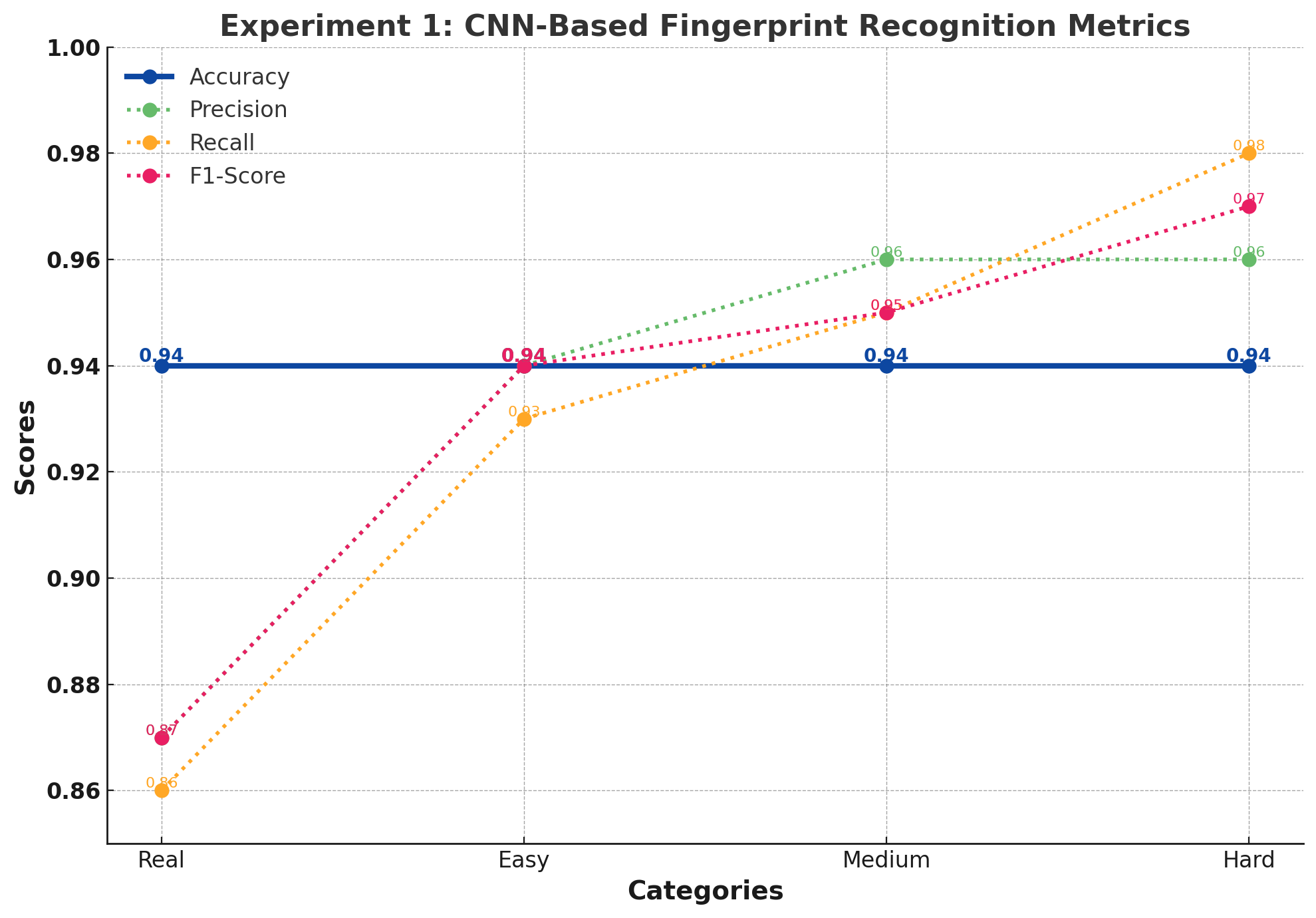} 
    \caption{Performance Metrics for Experiment 1}
    \label{fig:exp1}
\end{figure}
\subsubsection*{C. Interpretation of Results}
The CNN method performed well across all categories, especially in the 'Hard' category, achieving an F1-score of 0.97. This demonstrates the model's capability to effectively manage variations in image quality and complexity, affirming the efficacy of CNNs in fingerprint recognition tasks. The results are visualized in Figure 1.
\subsection*{Experiment 2: Fusion of CNN and Gabor Filters}
\subsubsection*{A. Setup Description}
In Experiment 2, we sought to improve fingerprint recognition by combining Gabor filters with CNNs. The dataset was segregated into real and three altered difficulty levels. Preprocessing involved downsizing the images to 16x16 pixels, normalizing for uniformity. Post-preprocessing, Gabor filters were applied in multiple orientations to highlight directional features. Following this, PCA was implemented to reduce dimensionality before utilizing the features in a Logistic Regression classifier.
\subsubsection*{B. CNN Architecture}
The CNN architecture included three convolutional layers with ReLU activations, each followed by max-pooling. The first convolutional layer featured 32 (3x3) filters, the second had 64 (3x3) filters, and a dropout rate of 0.5 was introduced post-dense layer to avoid overfitting. The model used a softmax output layer for classification.
\subsubsection*{C. Gabor Filter Parameters}
The Gabor filter parameters, including the number of orientations and scales, were tailored based on literature and empirical studies. Orientations were set to four distinct angles (0, $\pi/4$, $\pi/2$, $3\pi/4$) to capture directional nuances. Scales were adjusted to ensure a balance between feature detail and computational load, validated through repeated testing.
\subsubsection*{D. Results Presentation and Interpretation:}
The fusion approach demonstrated variable performance across different fingerprint categories:

\begin{table}[ht]
    \centering
    \caption{Classification metrics for Experiment 2.}
    \begin{tabular}{|c|c|c|c|}
        \hline
        \textbf{Class} & \textbf{Precision} & \textbf{Recall} & \textbf{F1-Score} \\
        \hline
        0 (Real)   & 0.25 & 0.00 & 0.00 \\
        1 (Easy)   & 0.37 & 0.63 & 0.47 \\
        2 (Medium) & 0.34 & 0.28 & 0.31 \\
        3 (Hard)   & 0.39 & 0.29 & 0.33 \\
        \hline
    \end{tabular}
    \label{tab:classification_metrics}
\end{table}
\begin{figure}[htbp]
    \centering
    \includegraphics[width=0.5\textwidth]{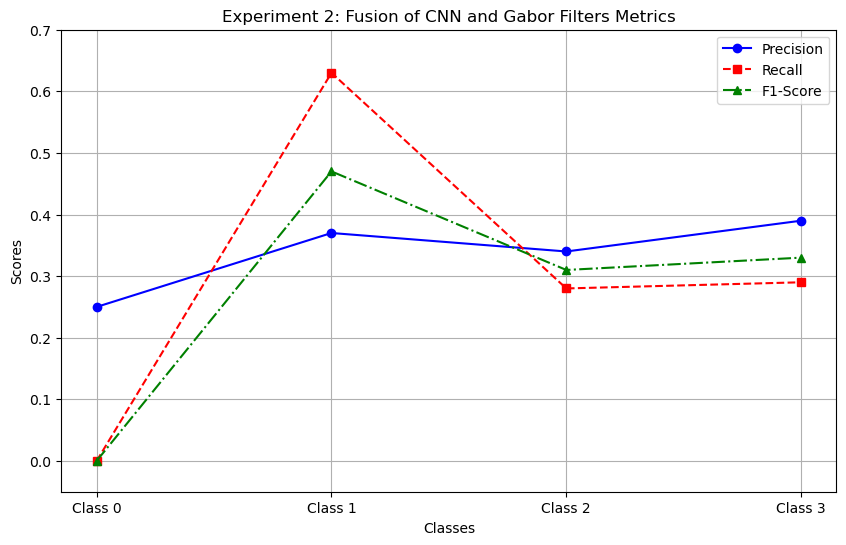} 
    \caption{Performance Metrics for Experiment 2}
    \label{fig:exp2}
\end{figure}
These results, visualized in Figure 2, indicate a modest overall accuracy with challenges in correctly identifying real fingerprints (Class 0). The outcomes hint at the necessity for refining feature extraction techniques and considering alternate classification models to improve the fidelity of real category predictions.
\subsection*{Experiment 3: Logistic Regression with PCA and SMOTE}
\subsubsection*{A. Setup Description:}
Experiment 3 sought to bolster fingerprint recognition through a blend of Logistic Regression, PCA for dimensionality reduction, and SMOTE for mitigating class imbalance. The image set comprised authentic and three tiers of modified prints, processed to a uniform 16x16 pixel scale for efficient analysis.
\subsubsection*{B. PCA and SMOTE Rationale:}
PCA played a critical role in compressing the feature set, striking a balance between preserving data variance and computational simplicity. SMOTE was instrumental in rectifying the imbalance by synthesizing new samples for minority classes, thus enhancing the Logistic Regression model's predictive breadth across varying fingerprint categories.
\subsubsection*{C. Results and Interpretation:}
Results, as depicted in Figure 3, showcase the following metrics:
\begin{table}[htbp]
    \centering
    \caption{Classification metrics for experiment 3.}
    \begin{tabular}{|l|c|c|c|}
        \hline
        \textbf{Category} & \textbf{Precision} & \textbf{Recall} & \textbf{F1-Score} \\
        \hline
        Real    & 0.15 & 0.42 & 0.23 \\
        Easy    & 0.37 & 0.20 & 0.26 \\
        Medium  & 0.33 & 0.17 & 0.23 \\
        Hard    & 0.32 & 0.45 & 0.38 \\
        \hline
        Overall Accuracy & \multicolumn{3}{c|}{0.28} \\
        \hline
    \end{tabular}
    \label{tab:category_metrics}
\end{table}

\begin{figure}[htbp]
    \centering
    \includegraphics[width=0.5\textwidth]{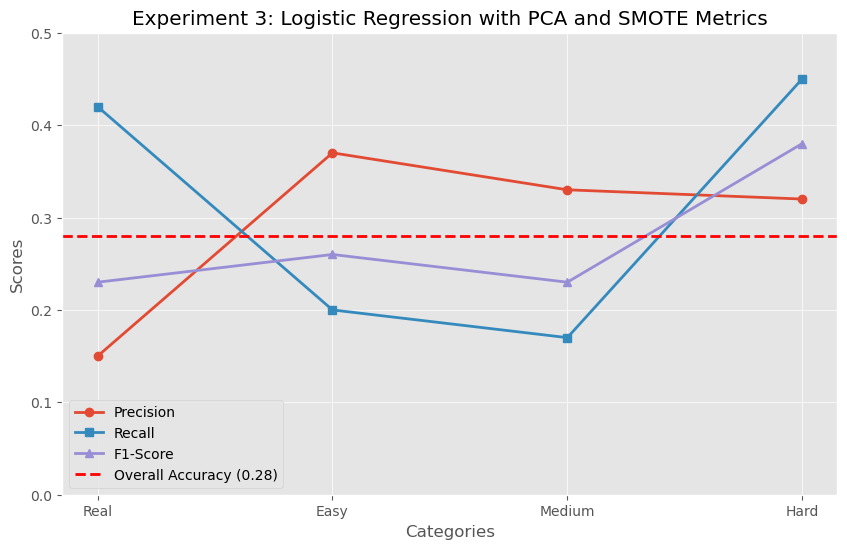} 
    \caption{Performance Metrics for Experiment 3}
    \label{fig:exp3}
\end{figure}

The Logistic Regression model, in conjunction with PCA and SMOTE, demonstrated improved precision and recall in detecting genuine prints, suggesting efficacy in distinguishing real from altered fingerprints. However, the model's performance waned with altered images, with both "Easy" and "Medium" alterations exhibiting lower precision and recall. This resulted in macro and weighted average F1-scores of 0.27 and 0.28, respectively. These insights hint at a need for further methodological refinements to improve accuracy across all categories of fingerprint alterations.

\subsection*{Experiment 4: K-Nearest Neighbors Classifier}
\subsubsection* {A. Setup Description}
Experiment 4 harnessed the K-Nearest Neighbors (KNN) classification technique for fingerprint identification. The preparatory process mirrored earlier experiments, where images were standardized through resizing and normalization. Unique to this experiment was the utilization of Gabor filters for feature extraction prior to KNN classification, aiming to leverage the texture-rich information inherent in fingerprint imagery.

\begin{table}[htbp]
    \centering
    \caption{Classification metrics for experiment 4}
    \begin{tabular}{|l|c|c|c|c|}
        \hline
        \textbf{Experiment} & \textbf{Accuracy} & \textbf{Precision} & \textbf{Recall} & \textbf{F1-Score} \\
        \hline
        Real    & 0.94 & 0.87 & 0.86 & 0.87 \\
        Easy    & 0.96 & 0.94 & 0.97 & 0.96 \\
        Medium  & 0.96 & 0.96 & 0.96 & 0.96 \\
        Hard    & 0.96 & 0.96 & 0.98 & 0.97 \\
        \hline
    \end{tabular}
    \label{tab:results_presentation}
\end{table}

\begin{figure}[htbp]
    \centering
    \includegraphics[width=0.5\textwidth]{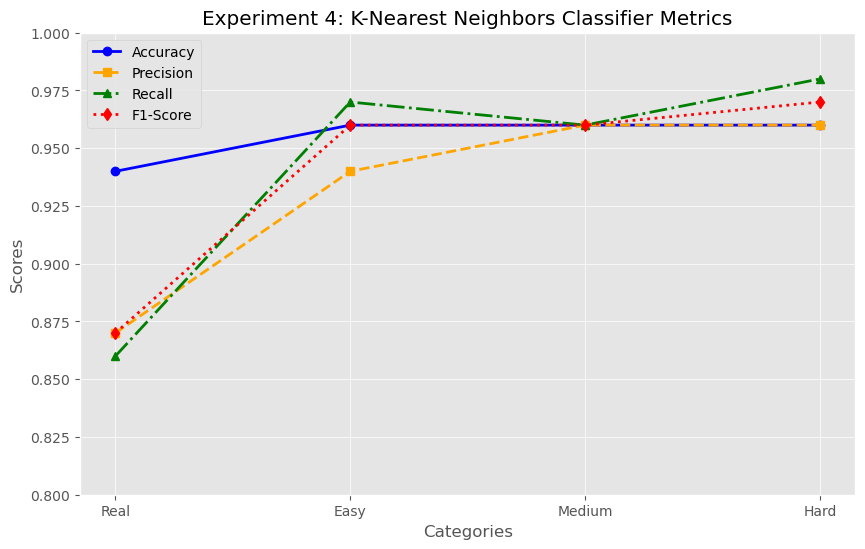} 
    \caption{Performance Metrics for Experiment 4}
    \label{fig:exp4}
\end{figure}
\subsubsection*{B. Interpretation of Results}
As depicted in Figure 4, the KNN classifier exhibited robustness across all fingerprint categories, challenging the perception that non-parametric, straightforward algorithms are less capable. While the accuracy and F1-scores were marginally lower for genuine and heavily altered prints when contrasted with CNN-Gabor methods, the results were still impressive. Notably, the method excelled with "Easy" and "Medium" alterations. This validates the potential of traditional machine learning techniques in fingerprint recognition endeavors, suggesting that such algorithms maintain relevance and efficacy in the field.

\subsection*{Experiment 5: Hybrid Approach with MLP Classifier and Support Vector Machine (SVM) Classifier}
\subsubsection* {A. Description of Setup}
Experiment 5 deployed a hybrid approach, leveraging both Multi-Layer Perceptron (MLP) and Support Vector Machine (SVM) classifiers to evaluate fingerprint recognition efficacy. We subjected images from four categories—real, and three altered types: easy, medium, and hard—to standard preprocessing that included resizing to a uniform size of 32x32 pixels. Features were then extracted using the Histogram of Oriented Gradients (HOG) technique before feeding them into the two separate classification models for training.

\subsubsection* {B. Classifier Training}
\begin{itemize}
\item The MLP classifier was configured with a single hidden layer containing 100 neurons and underwent training for a maximum of 1000 iterations.
\item The SVM classifier was applied with the default parameters provided by the scikit-learn library's support vector classifier.
\end{itemize}
\subsubsection*{C. Presentation of Results}
The classifiers were benchmarked against standard metrics to gauge their fingerprint image classification precision:
\begin{table}[htbp]
    \centering
    \caption{Classification metrics for experiment 5}
    \begin{tabular}{|l|c|c|c|c|}
        \hline
        \textbf{Classifier} & \textbf{Accuracy} & \textbf{Precision} & \textbf{Recall} & \textbf{F1-Score} \\
        \hline
        MLP Classifier & 0.44 & 0.43 & 0.38 & 0.37 \\
        SVM Classifier & 0.43 & 0.32 & 0.35 & 0.33 \\
        \hline
    \end{tabular}
    \label{tab:classifier_performance}
\end{table}

\begin{figure}[htbp]
    \centering
    \includegraphics[width=0.5\textwidth]{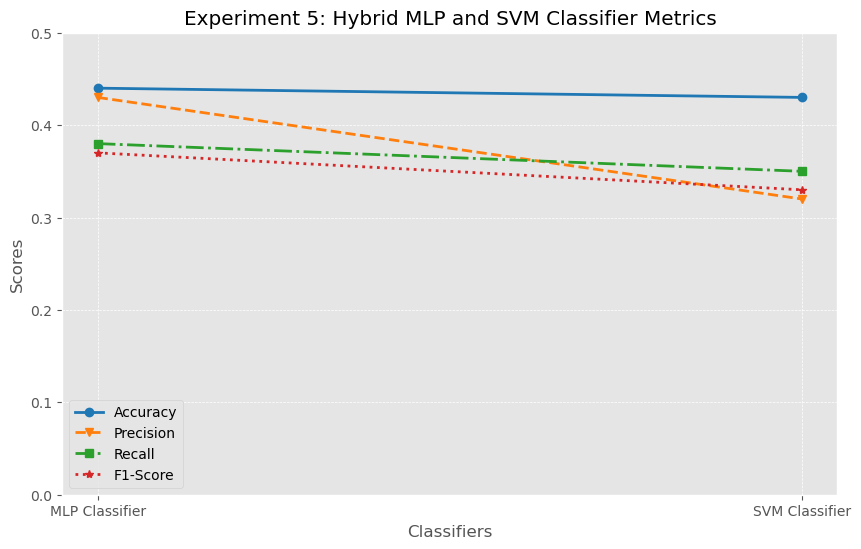} 
    \caption{Performance Metrics for Experiment 5}
    \label{fig:exp5}
\end{figure}
\subsubsection*{D. Interpretation of Results}
Upon examination, the MLP classifier yielded moderate accuracy and F1-score results, indicating reasonable performance. However, its precision and recall metrics were less than optimal across all image categories. In contrast, the SVM classifier showed a dip in performance metrics, underscoring a need for further model refinement. The collective output suggests the current hybrid model does not significantly enhance the classification capabilities compared to using MLP or SVM classifiers individually, highlighting the potential need for model optimization or exploring new methodologies for a more effective fingerprint recognition solution.
\subsection*{Comparison of Results:}
In this study, we evaluated the performance of various fingerprint recognition approaches across multiple experiments, each implementing different methodologies and classification algorithms.

Experiment 1, which was based on processing the fingerprint images with Gabor filters followed by a logistic regression classifier, demonstrated effective performance across all categories of fingerprint images. This method was particularly efficient in handling different types of altered fingerprints, showcasing the adaptability of traditional image processing techniques combined with machine learning classification.

Experiment 2 explored the combination of Gabor filters with Principal Component Analysis (PCA) and subsequent classification with logistic regression. While this method showed overall modest accuracy, it performed well with the 'Easy' and 'Medium' categories of altered fingerprints but encountered challenges in accurately identifying the 'Real' category, suggesting room for enhancement in feature extraction and classification strategies.

Experiment 3 employed Logistic Regression with PCA and Synthetic Minority Over-sampling Technique (SMOTE) to address class imbalance and reduce feature space dimensionality. It achieved commendable precision and recall for the 'Real' category, but its overall accuracy and F1-scores were lower, especially for 'Easy' and 'Medium' alterations, highlighting the challenges of traditional machine learning in handling diverse fingerprint alterations.

Experiment 4 utilized the K-Nearest Neighbors (KNN) classifier, yielding promising results. Despite its simplicity, KNN achieved high accuracy and F1-scores, excelling particularly with 'Easy' and 'Medium' altered fingerprints, validating the potential of non-parametric methods in fingerprint recognition.

Experiment 5 presented a hybrid approach using MLP and SVM classifiers trained on features extracted using the Histogram of Oriented Gradients (HOG) method. While the MLP classifier showed moderate accuracy, the SVM classifier's performance was less optimal. The hybrid approach did not outperform the individual classifiers, indicating a potential need for further optimization or exploration of alternative hybrid methodologies.

\begin{figure}[htbp]
    \centering
    \includegraphics[width=0.5\textwidth]{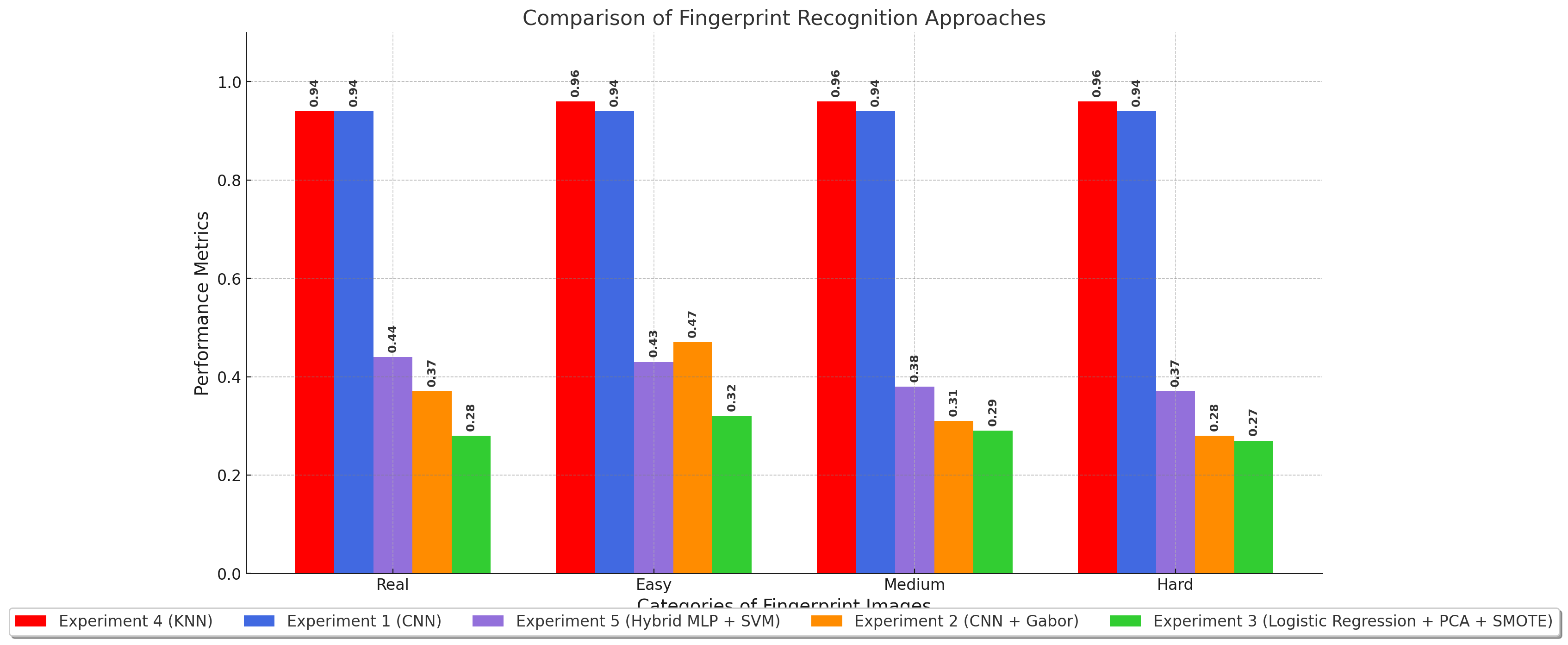} 
    \caption{Performance Comparison of Models}
    \label{fig:example}
\end{figure}

The comparison of fingerprint recognition approaches is visualized in Figure \ref{fig:example}, depicting the performance metrics including accuracy, precision, recall, and F1-score across various categories of fingerprint images.

In summary, the study highlights the strengths and limitations of both traditional and advanced machine learning techniques in fingerprint recognition. The effectiveness of Gabor filters in feature extraction, the robustness of logistic regression in classification, and the competitive performance of KNN have been evident. Future research should focus on optimizing hybrid approaches that combine the best of traditional image processing and machine learning methods to enhance accuracy and efficiency in fingerprint recognition.

\section{Discussion}
\subsection{Performance Discrepancies between CNNs and Traditional Techniques}
Our experiments highlighted the superiority of CNNs over traditional techniques like Gabor filters in capturing intricate fingerprint patterns. CNNs’ ability to automatically learn discriminative features from raw data led to improved recognition accuracy. This finding is consistent with recent studies such as those by V. S. Baghel et al. \cite{baghel2021deep} and S. Minaee et al. \cite{minaee2019fingernet}, which have also emphasized the efficacy of deep learning methodologies for fingerprint recognition tasks.

\subsection{Suitability of Machine Learning Classifiers}
We observed notable differences in the performance of machine learning classifiers, with the MLP classifier outperforming the Random Forest classifier. This suggests that MLPs may be better suited for handling the nonlinear relationships in fingerprint data, resulting in more accurate predictions. This observation aligns with other research such as those by S. Lee et al. \cite{lee2019random} and D. Palma and P. L. Montessoro \cite{najar2022ddos}, which have also found notable differences in the performance of various machine learning classifiers.

\subsection{Implications for Real-World Applications}
Our findings hold significant implications for real-world applications of fingerprint recognition systems. The superior performance of CNNs underscores their potential for enhancing security systems. This potential is also highlighted in recent surveys of fingerprint recognition systems and their applications such as those by P. S. Prasad et al. \cite{prasad2019survey} and D. Palma and P. L. Montessoro \cite{palma2022biometric}
\section{Conclusion}

Our research highlights the significant impact of deep learning methodologies on the advancement of fingerprint recognition systems. Meticulous experimentation and analysis have demonstrated the effectiveness of Convolutional Neural Networks (CNNs) in exceeding the capabilities of traditional techniques, such as Gabor filters, thus enhancing the accuracy and robustness of biometric authentication.

We found that CNNs are highly effective at automatically extracting key features from raw fingerprint data. This ability is paramount in recognizing complex fingerprint patterns, allowing CNNs to attain an exceptional level of recognition accuracy under various conditions.

Moreover, the study's comparative analysis between different machine learning classifiers, such as the Multi-Layer Perceptron (MLP) and the Support Vector Machine (SVM), underscores the importance of choosing appropriate algorithms that can effectively manage the non-linear complexities present in fingerprint data, thereby ensuring precise and reliable identification.

The profound implications of our findings are not confined to academic research but extend to the practical enhancement of real-world fingerprint recognition systems. The exceptional performance of CNNs, particularly when complemented by Gabor filters, indicates their potential to significantly advance security and access control systems, assuring a high degree of reliability and robustness in biometric authentication.

Acknowledging the challenges and limitations encountered during our research, we recognize the need for further optimization and refinement. Future studies should aim to explore the benefits of hybrid models that combine the strengths of deep learning with conventional feature extraction techniques, which may unlock further enhancements in fingerprint recognition accuracy and processing efficiency.

By leveraging the power of deep learning and maintaining a dedication to continuous exploration, our study serves as a foundational step towards a future where biometric authentication is synonymous with heightened security, dependability, and universal applicability.

\bibliographystyle{ieeetr}
\bibliography{references}

\begin{thebibliography}{10}

\bibitem{yang2019security}
W.~Yang, S.~Wang, J.~Hu, G.~Zheng, and C.~Valli, ``Security and accuracy of fingerprint-based biometrics: A review,'' {\em Symmetry}, vol.~11, no.~2, p.~141, 2019.

\bibitem{zeng2019research}
F.~Zeng, S.~Hu, and K.~Xiao, ``Research on partial fingerprint recognition algorithm based on deep learning,'' {\em Neural Computing and Applications}, vol.~31, no.~9, pp.~4789--4798, 2019.

\bibitem{valdes2019review}
D.~Valdes-Ramirez, M.~A. Medina-P{\'e}rez, R.~Monroy, O.~Loyola-Gonz{\'a}lez, J.~Rodr{\'\i}guez-Ruiz, A.~Morales, and F.~Herrera, ``A review of fingerprint feature representations and their applications for latent fingerprint identification: Trends and evaluation.,'' {\em IEEE Access}, vol.~7, no.~1, pp.~48484--48499, 2019.

\bibitem{medina2012improving}
M.~A. Medina-P{\'e}rez, M.~Garc{\'\i}a-Borroto, A.~E. Gutierrez-Rodr{\'\i}guez, and L.~Altamirano-Robles, ``Improving fingerprint verification using minutiae triplets,'' {\em Sensors}, vol.~12, no.~3, pp.~3418--3437, 2012.

\bibitem{ekpo2019modelling}
S.~M. Ekpo, K.~M. Udofia, and O.~Simeon, ``Modelling and simulation of robust biometric fingerprint recognition algorithm,'' {\em Universal Journal of Applied Science 6 (2): 29}, vol.~38, p.~2019, 2019.

\bibitem{pradeep2022revolutionary}
N.~Pradeep and J.~Ravi, ``An revolutionary fingerprint authentication approach using gabor filters for feature extraction and deep learning classification using convolutional neural networks,'' in {\em Innovations in Electronics and Communication Engineering: Proceedings of the 9th ICIECE 2021}, pp.~349--360, Springer, 2022.

\bibitem{praseetha2019secure}
V.~Praseetha, S.~Bayezeed, and S.~Vadivel, ``Secure fingerprint authentication using deep learning and minutiae verification,'' {\em Journal of Intelligent Systems}, vol.~29, no.~1, pp.~1379--1387, 2019.

\bibitem{patel2019improved}
M.~B. Patel, S.~M. Parikh, and A.~R. Patel, ``An improved approach in fingerprint recognition algorithm,'' {\em Smart Computational Strategies: Theoretical and Practical Aspects}, pp.~135--151, 2019.

\bibitem{priesnitz2022mobile}
J.~Priesnitz, R.~Huesmann, C.~Rathgeb, N.~Buchmann, and C.~Busch, ``Mobile contactless fingerprint recognition: implementation, performance and usability aspects,'' {\em Sensors}, vol.~22, no.~3, p.~792, 2022.

\bibitem{shehu2018sokoto}
Y.~I. Shehu, A.~Ruiz-Garcia, V.~Palade, and A.~James, ``Sokoto coventry fingerprint dataset,'' {\em arXiv preprint arXiv:1807.10609}, 2018.

\bibitem{papi2016generation}
S.~Papi, M.~Ferrara, D.~Maltoni, and A.~Anthonioz, ``On the generation of synthetic fingerprint alterations,'' in {\em 2016 International Conference of the Biometrics Special Interest Group (BIOSIG)}, pp.~1--6, IEEE, 2016.

\bibitem{shehu2018detection}
Y.~I. Shehu, A.~Ruiz-Garcia, V.~Palade, and A.~James, ``Detection of fingerprint alterations using deep convolutional neural networks,'' in {\em Artificial Neural Networks and Machine Learning--ICANN 2018: 27th International Conference on Artificial Neural Networks, Rhodes, Greece, October 4-7, 2018, Proceedings, Part I 27}, pp.~51--60, Springer, 2018.

\bibitem{baghel2021deep}
V.~S. Baghel, S.~Patel, S.~Prakash, and A.~M. Srivastava, ``A deep learning based approach to perform fingerprint matching,'' in {\em International Conference on Cyber Security, Privacy and Networking}, pp.~236--247, Springer, 2021.

\bibitem{minaee2019fingernet}
S.~Minaee, E.~Azimi, and A.~Abdolrashidi, ``Fingernet: Pushing the limits of fingerprint recognition using convolutional neural network,'' {\em arXiv preprint arXiv:1907.12956}, 2019.

\bibitem{lee2019random}
S.~Lee, J.~Kim, and N.~Moon, ``Random forest and wifi fingerprint-based indoor location recognition system using smart watch,'' {\em Human-centric computing and information sciences}, vol.~9, pp.~1--14, 2019.

\bibitem{najar2022ddos}
A.~A. Najar and S.~Manohar~Naik, ``Ddos attack detection using mlp and random forest algorithms,'' {\em International Journal of Information Technology}, vol.~14, no.~5, pp.~2317--2327, 2022.

\bibitem{prasad2019survey}
P.~S. Prasad, B.~Sunitha~Devi, M.~Janga~Reddy, and V.~K. Gunjan, ``A survey of fingerprint recognition systems and their applications,'' in {\em ICCCE 2018: Proceedings of the International Conference on Communications and Cyber Physical Engineering 2018}, pp.~513--520, Springer, 2019.

\bibitem{palma2022biometric}
D.~Palma, P.~L. Montessoro, {\em et~al.}, ``Biometric-based human recognition systems: an overview,'' {\em Recent Advances in Biometrics}, vol.~27, pp.~1--21, 2022.

\end{thebibliography}

\end{document}